\documentclass[conference]{IEEEtran}
\IEEEoverridecommandlockouts
\usepackage{cite}
\usepackage{amsmath,amssymb,amsfonts}
\usepackage{algorithmic}
\usepackage{graphicx}
\usepackage{textcomp}
\usepackage{xcolor}
\usepackage{url}
\usepackage{hyperref}
\usepackage{float}
\usepackage{booktabs}
\def\BibTeX{{\rm B\kern-.05em{\sc i\kern-.025em b}\kern-.08em
    T\kern-.1667em\lower.7ex\hbox{E}\kern-.125emX}}

\begin{document}

\title{Memory-First Fact-Checking: A Knowledge-Graph-Grounded Multi-Agent System for Misinformation Detection}

\author{\IEEEauthorblockN{Amelia Petrenciuc, Alexandru Lecu and Adrian Groza}
 \IEEEauthorblockA{ 
 \textit{Artificial Intelligence Research Institute AIRi@UTCN}\\
 \textit{Technical University of Cluj-Napoca, Cluj-Napoca, Romania}\\
 \texttt{amelia.petenciuc@campus.utcluj.ro, adrian.groza@cs.utcluj.ro}
 }
}

\maketitle

\begin{abstract}
This paper introduces a hybrid fact-checking framework that integrates Knowledge Graph-based semantic memory with adversarial multi-agent reasoning for explainable misinformation detection. The proposed system follows a \textit{memory-first, web-fallback} architecture, in which input claims are initially evaluated against a dual-index Knowledge Graph through Sentence-BERT-based semantic retrieval and Natural Language Inference. When the evidence retrieved from the graph is insufficient to support a reliable decision, the framework collects information from trusted web sources and assesses it using an adversarial tribunal composed of support, contradiction, and judging agents. A graph-aware confidence mechanism combines semantic similarity, NLI confidence, and structural graph evidence to determine whether internal knowledge is sufficient, thereby reducing unnecessary web retrieval. Following verification, validated information is transformed into structured triples and incorporated into the Knowledge Graph, supporting the incremental expansion of the system's semantic memory. Experimental evaluation on a curated COVID-19 misinformation benchmark demonstrates that the proposed framework achieves an accuracy of 97.4\% and a macro-averaged F1-score of 92.6\% on resolved claims, outperforming a Llama~3.3~70B baseline, which obtains an accuracy of 87.7\% and a macro-averaged F1-score of 86.3\%.
\end{abstract}


\section{Introduction}

The rapid dissemination of misinformation across digital media has increased the need for automated fact-checking methods capable of operating at scale~\cite{guo2022}. The COVID-19 pandemic further illustrated this challenge, as large volumes of information originating from both reliable and questionable sources circulated across multiple social media platforms~\cite{cinelli2020}.

Existing automated approaches nevertheless exhibit several important limitations. Content-based classifiers and fine-tuned transformer models typically infer veracity from the linguistic characteristics of an input claim or article~\cite{shu2017,kaliyar2021}. Although such models can achieve strong classification performance, they do not inherently retrieve external evidence that can be independently inspected by the user. Large language models (LLMs) introduce an additional reliability concern, as their generated outputs may contain plausible but unsupported information, commonly characterized as hallucination~\cite{ji2023}. Multi-agent debate has been shown to improve factual validity and reasoning by enabling multiple model instances to examine and challenge one another's responses~\cite{du2024}.

Knowledge Graphs (KGs) provide structured and reusable representations of factual information; however, a preconstructed graph cannot verify claims that are not covered by its stored knowledge. Retrieval-augmented generation (RAG) addresses part of this limitation by providing language models with access to explicit non-parametric memory and retrieved source passages~\cite{lewis2020}. Nevertheless, fact verification requires more than identifying semantically relevant evidence: the retrieved information must also be assessed to determine whether it supports, refutes, or provides insufficient information for the claim under analysis~\cite{thorne2018fever}.

To address these complementary limitations, this paper introduces a hybrid fact-checking framework with four main contributions: (i)~a \textit{memory-first, web-fallback} architecture that initially evaluates claims against a dual-index KG and invokes external evidence retrieval only when the internal knowledge is insufficient; (ii)~a graph-aware confidence scoring mechanism that combines semantic similarity, Natural Language Inference (NLI) confidence, and structural graph support to determine whether web retrieval is required; (iii)~an adversarial deliberation tribunal composed of support, contradiction, and judging agents that independently evaluates competing interpretations of the retrieved evidence; and (iv)~a persistent knowledge enrichment mechanism that transforms newly verified web evidence into structured triples and incorporates them into the KG for subsequent reuse.

\section{Related Work}
\label{sec:related-work}

\subsection{Knowledge-Grounded Fact Verification}

Content-based misinformation detection has evolved from classifiers based on linguistic and propagation features~\cite{shu2017} to transformer-based architectures such as FakeBERT~\cite{kaliyar2021}. Although these methods can identify predictive textual patterns, they do not inherently provide external evidence that can be independently inspected.

Evidence-grounded approaches instead verify claims against external knowledge \cite{DBLP:journals/corr/abs-2004-12330}. FEVER established a widely used formulation in which claims are classified as \textsc{Supported}, \textsc{Refuted}, or \textsc{Not Enough Info} based on retrieved textual evidence~\cite{thorne2018fever}.

Knowledge-graph-based verification represents evidence through entities
and relations, while FactKG evaluates structured reasoning over such
graphs~\cite{kim2023factkg}. However, the effectiveness of such approaches remains constrained by the coverage of the underlying graph, particularly for emerging or domain-specific claims \cite{cheres2025mindbugs}.

To improve coverage, Kolli \textit{et al.} combine DBpedia-based retrieval, language-model classification, and a web-search agent that is invoked when the KG does not provide sufficient evidence~\cite{kolli2025}. Their work establishes the relevance of hybrid KG--web verification. Building on this direction, the framework proposed in this paper focuses on dual-index semantic memory, graph-aware confidence estimation, explicit support--contradiction deliberation, and the persistent integration of newly verified evidence into the KG.

\subsection{Semantic Retrieval and Stance-Aware Inference}

Dense retrieval based on Sentence-BERT (SBERT) enables efficient matching between semantically related claims and evidence passages~\cite{reimers2019}. Retrieval-augmented generation (RAG) similarly grounds language-model outputs in external, non-parametric knowledge~\cite{lewis2020}. However, retrieval relevance alone does not determine the evidential stance of a passage: semantically similar statements may support, refute, or remain neutral with respect to the same claim.

Natural Language Inference (NLI) provides this complementary distinction by modeling entailment, contradiction, and neutrality between a premise and a hypothesis~\cite{williams2018}. Nevertheless, NLI models may rely on lexical overlap and remain vulnerable to negation and numerical reasoning errors~\cite{naik2018}. These limitations are particularly relevant to fact-checking, where small changes in quantities, dates, or qualifiers may reverse the meaning of a claim.

The proposed framework therefore separates evidence retrieval from stance assessment. Claim-level and atomic-fact-level indexes identify candidate evidence, while NLI evaluates its logical relationship with the input claim. Graph structure and safeguards for numerically sensitive inferences are then used to determine whether the KG evidence is sufficiently reliable or whether external verification is required.

\subsection{Multi-Agent Reasoning and Persistent Knowledge Enrichment}

Multi-agent debate has been investigated as a means of improving language-model reasoning by allowing multiple model instances to challenge and revise candidate responses. Du \textit{et al.} employ iterative debate among model instances to improve factuality and reasoning reliability~\cite{du2024}. Liang \textit{et al.} similarly introduce opposing agents and a judging agent to promote divergent reasoning and reduce commitment to an initially generated solution~\cite{liang2024}. Avram \textit{et al.} orchestrate multiple heterogeneous agents through the Model Context Protocol for automated disinformation detection~\cite{avram2025}.

A related direction addresses complex fact verification through explicit task decomposition. Program-Guided Fact-Checking (ProgramFC) generates reasoning programs that decompose complex claims into simpler subtasks executed through specialized functions~\cite{pan2023}. Although ProgramFC is not an adversarial multi-agent architecture, it demonstrates the value of separating verification into explicit and interpretable reasoning operations.

The proposed framework combines these principles through specialized support, contradiction, and judging roles. Unlike general-purpose debate, the agents evaluate competing interpretations of the same retrieved evidence, and the final decision is grounded in the cited passages rather than in unconstrained model interaction.

Persistent KG enrichment additionally requires entity resolution to prevent equivalent mentions from being stored as separate nodes. Existing methods range from normalization and string-based record matching~\cite{christen2012} to neural retrieval of canonical entities~\cite{decao2021}. The proposed system adopts a lightweight resolution strategy tailored to incremental graph updates, enabling newly verified facts and aliases to be reused during subsequent verification tasks.

\section{System Architecture}
\label{sec:architecture}

\subsection{System Overview}
\label{subsec:system-overview}

\begin{figure*}
\centering
\includegraphics[width=0.85\textwidth]{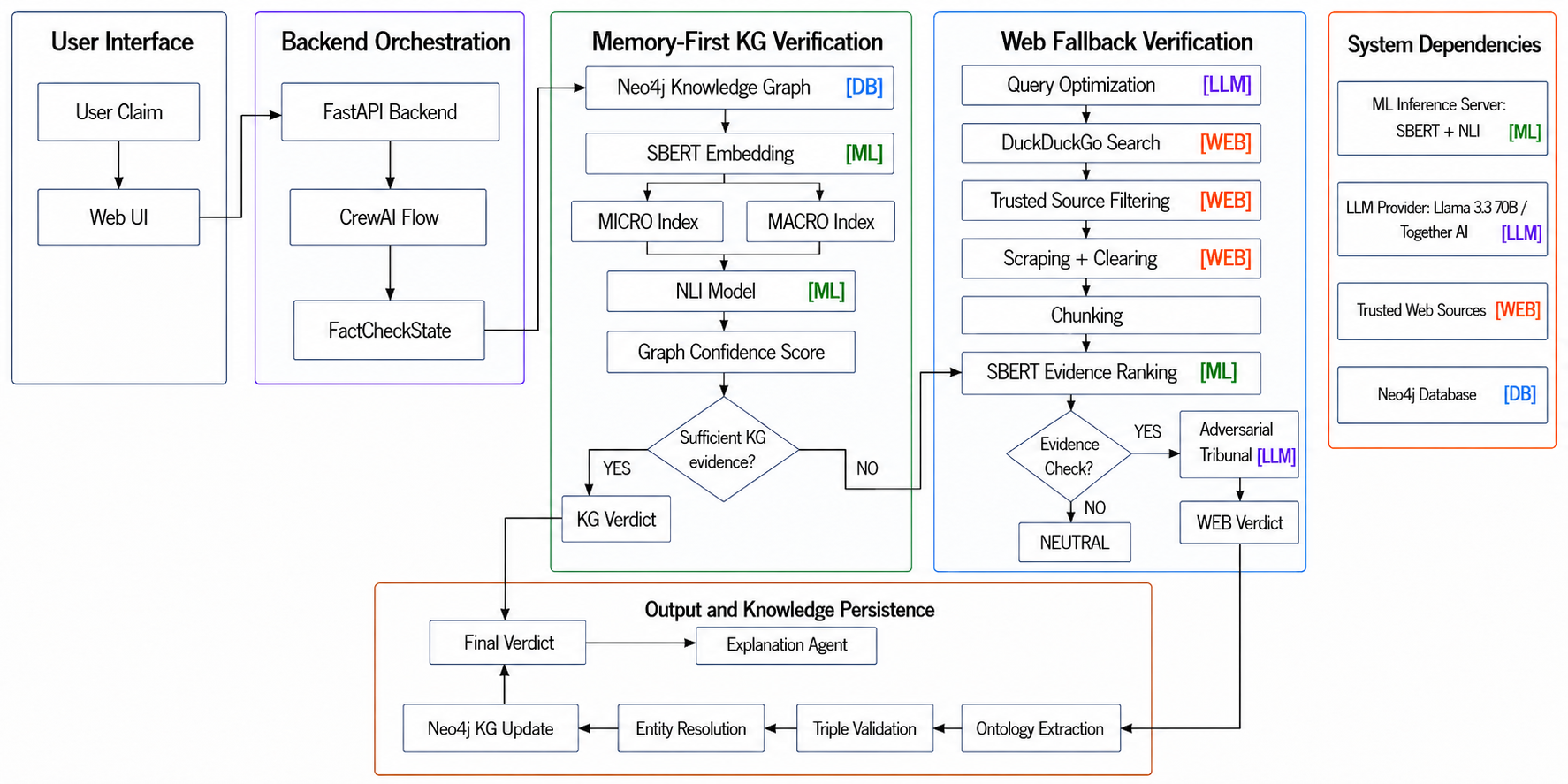}
\caption{Overall architecture of the memory-first, web-fallback fact-checking framework.}
\label{fig:architecture}
\end{figure*}

The proposed framework follows a \textit{memory-first, web-fallback} architecture, illustrated in Fig.~\ref{fig:architecture}. Given an input claim, the system first queries a persistent Knowledge Graph (KG) containing previously evaluated claims and atomic facts. The retrieved graph candidate is assessed through semantic similarity, Natural Language Inference (NLI), and graph-aware confidence scoring. Claims that cannot be resolved reliably from internal memory are transferred to a web evidence pipeline and subsequently evaluated by an adversarial multi-agent tribunal.

Verified conclusions obtained through external investigation are transformed into structured Subject--Predicate--Object facts and persisted in Neo4j. This creates a feedback loop in which a claim that initially requires web verification may later be resolved from accumulated graph memory. The workflow is orchestrated through CrewAI, while embedding generation and NLI inference are executed by a dedicated FastAPI service.

Verdict generation is separated from user-facing explanation generation. The factual label is established either by the deterministic KG decision layer or by the evidence-grounded tribunal. Downstream explanation components only summarize the evidence associated with the established result and cannot modify the verdict.
\subsection{Dual-Index Knowledge Graph}
\label{subsec:dual-index-kg}

The Knowledge Graph represents the persistent semantic memory of the system and maintains two complementary vector indexes. The \textbf{MACRO index} operates on complete \texttt{Claim} nodes and preserves their contextual formulation. The \textbf{MICRO index} operates on atomic \texttt{VerifiedFact} and \texttt{ClaimedFact} nodes represented as Subject--Predicate--Object statements. Both indexes use 768-dimensional embeddings generated with \texttt{all-mpnet-base-v2} and cosine similarity for retrieval.

For an input claim $c$, the system retrieves the highest-scoring candidate from each index. 
A graph node $n$ is considered eligible only when
\begin{equation}
S(c,n) \geq 0.75,
\label{eq:candidate-threshold}
\end{equation}
where $S(c,n)$ denotes the cosine similarity between the input claim and the indexed node. If neither index returns an eligible candidate, the claim is routed directly to external verification.

When both indexes return eligible candidates, a deterministic arbitration rule selects the appropriate representation level. For claims containing at most six words, the MICRO candidate is preferred when the difference between the two similarity scores does not exceed 0.03. Longer claims generally favor the MACRO candidate because it preserves more contextual information. However, the MICRO result overrides this preference when its similarity exceeds the MACRO score by more than 0.05. In all remaining cases, the candidate with the higher similarity score is selected.

\subsection{NLI and Graph-Aware Decision Layer}
\label{subsec:nli-layer}

The selected candidate is evaluated using the
\texttt{facebook/bart-large-mnli} model. The retrieved graph statement is used as the premise, while the input claim is used as the hypothesis. The model estimates the probabilities of the \textsc{Entailment}, \textsc{Contradiction}, and \textsc{Neutral} relations.

The predicted NLI relation is not treated as a factual verdict in isolation. Instead, it is interpreted jointly with the truth status stored in the retrieved graph node, according to Table~\ref{tab:nli-truth-table}. Entailment with a verified true statement supports the input claim, whereas entailment with a stored false statement indicates that the input reproduces previously identified misinformation. Contradiction reverses the stored interpretation subject to semantic reliability safeguards.

\begin{table}
\centering
\caption{Truth-aware interpretation of the NLI relation.}
\label{tab:nli-truth-table}
\setlength{\tabcolsep}{4pt}
\small
\begin{tabular}{lll}
\toprule
\textbf{Stored status} &
\textbf{NLI relation} &
\textbf{Verdict}
\tabularnewline
\midrule
TRUE  & Entailment    & TRUE
\tabularnewline
TRUE  & Contradiction & FALSE
\tabularnewline
FALSE & Entailment    & FALSE
\tabularnewline
FALSE & Contradiction & TRUE$^{*}$
\tabularnewline
Any   & Neutral       & Confidence-gated
\tabularnewline
\bottomrule
\end{tabular}

\vspace{1mm}
\begin{minipage}{0.94\columnwidth}
\footnotesize
$^{*}$Accepted only when corrective evidence is available in the graph neighborhood.
\end{minipage}
\end{table}

To estimate whether internal evidence is sufficiently reliable, the system computes a graph-aware confidence score:
\begin{equation}
C = 0.60*S + 0.30*N + 0.10*G,
\label{eq:confidence}
\end{equation}
where $S$ is the semantic similarity score, $N$ is the directional confidence of the NLI, and $G$ represents structural support from the graph neighborhood. The weights were set empirically and kept fixed for evaluation:
semantic similarity is dominant because candidate relevance is a
prerequisite for a reliable decision, NLI determines evidential
stance, and graph structure provides secondary corroboration. The directional NLI component is defined as
\begin{equation}
N =
\max \left(
P(\mathrm{ENTAILMENT}),
P(\mathrm{CONTRADICTION})
\right).
\label{eq:nli-confidence}
\end{equation}
The graph-support component captures corroborating facts, connected entities, and corrective relations associated with the selected node. The resulting composite value is used as a routing signal rather than as a calibrated probability. Neutral NLI outcomes require stronger semantic and structural evidence than directional outcomes before a graph verdict can be accepted.

Additional safeguards prevent weak contradictions from reversing stored knowledge. Path-specific similarity checks are applied separately to MACRO and MICRO candidates. A numeric guard detects pairs with similar linguistic structure but different numerical, temporal, or regional values, while a high-neutrality guard rejects matches for which the NLI model expresses strong logical uncertainty. Whenever a safeguard is activated, the internal result is treated as unresolved and the claim is transferred to web verification. 

\subsection{Web Evidence Pipeline}
\label{subsec:web-evidence-pipeline}

The web branch is activated when no eligible graph candidate is available, when the graph-aware confidence is insufficient, or when an NLI safeguard rejects the retrieved relation. Evidence collection is completed before any deliberative agent is invoked.

An LLM-assisted query optimizer transforms the input claim into a concise search query while preserving the main entities, factual relation, polarity, and relevant numerical information. The optimized query is submitted to DuckDuckGo, which returns up to 40 candidate results. Retrieved URLs are filtered through a curated whitelist containing fact-checking organizations, public-health institutions, scientific and medical sources, and established news agencies.

Up to five pages from distinct trusted domains are scraped and cleaned. The extracted text is divided into passages of 100 words with an overlap of 25 words. Each passage retains its source URL and is ranked against the original claim using \texttt{all-mpnet-base-v2} embeddings. Passages containing explicit fact-checking or refutation language receive small relevance adjustments.

The 15 highest-ranked passages form an intermediate candidate pool. A domain-aware diversification step then retains at most two passages from the same domain. The final evidence bundle contains up to six passages, each associated with its relevance score and source URL. If the highest passage similarity remains below 0.40, the available evidence is considered insufficient and the system returns \textsc{Neutral} without invoking the tribunal.

\subsection{Adversarial Deliberation Tribunal}
\label{subsec:adversarial-tribunal}

\begin{figure}
\centering
\includegraphics[width=\columnwidth]{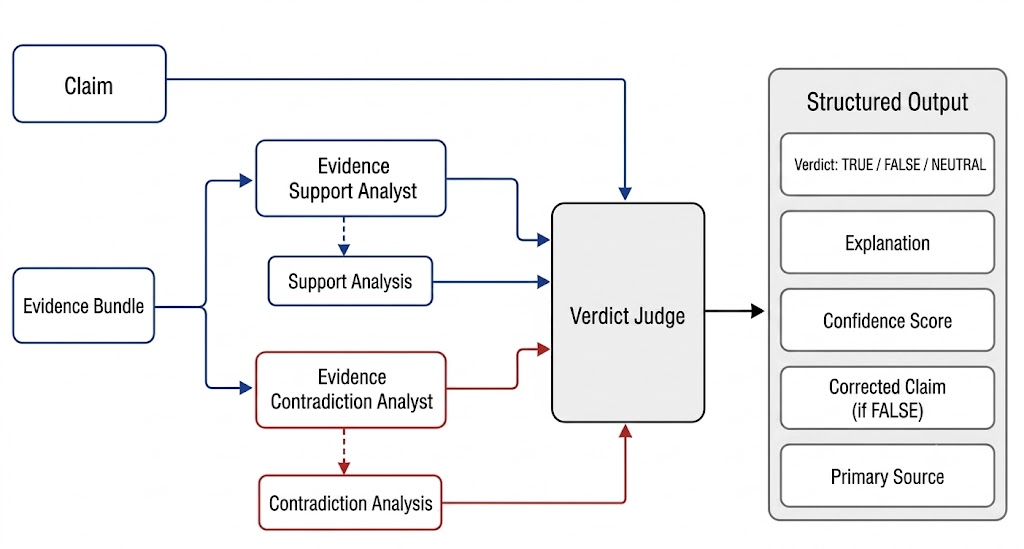}
\caption{Evidence and analysis flow within the adversarial multi-agent tribunal.}
\label{fig:tribunal}
\end{figure}

Evidence bundles that pass the sufficiency check are evaluated by a three-agent tribunal coordinated through a sequential CrewAI process. All agents use \texttt{Llama-3.3-70B-Instruct-Turbo} through Together AI, with temperature set to 0.0. The information flow between the agents is illustrated in Fig.~\ref{fig:tribunal}.

The \textbf{Support Analyst} receives the claim and the complete evidence bundle and identifies passages that directly or partially support the assertion. The \textbf{Contradiction Analyst} examines the same inputs for direct contradictions, narrowing conditions, missing context, and unsupported generalizations. Both analysts are restricted to the retrieved evidence and must associate their conclusions with source URLs.

The \textbf{Judge} receives the original claim, the complete evidence bundle, and both analyses. It returns a structured verdict in
$\{\textsc{True},\textsc{False},\textsc{Neutral}\}$, together with an explanation, confidence estimate, primary source, and a corrected assertion when the verdict is \textsc{False}. The Judge evaluates evidence quality, specificity, and coverage rather than the rhetorical strength of the analyst outputs. When the evidence remains insufficient or contradictory, \textsc{Neutral} is preserved as a valid abstention.

A single-pass web-verification routine is used only when tribunal execution fails technically, for example because of an API error or an invalid model response. It is not activated when the Judge returns a valid \textsc{Neutral} verdict.

\subsection{Knowledge Persistence}
\label{subsec:knowledge-persistence}

After a successful web verification, an ontology extraction component transforms the claim and its supporting evidence into structured Subject--Predicate--Object triples. The extraction distinguishes between the content asserted by the original claim and the factual statements supported by the retrieved evidence. Triples containing vague entities, malformed predicates, placeholder values, or discourse-level expressions are rejected before insertion.

Entities and predicates are normalized prior to persistence. Exact normalized matching, predefined aliases, and fuzzy lexical matching are used to resolve new surface forms against entities already present in Neo4j. Verified facts are embedded with the same sentence encoder used by the retrieval layer, enabling future MICRO-level semantic search.

Persistence depends on the final verdict. For a \textsc{True} result, the verified claim and its supporting facts are stored in the graph. For a \textsc{False} result, the original misinformation claim may be stored together with a corrected verified claim, linked through a \texttt{CORRECTED\_BY} relation. Neutral claims are not persisted because they do not represent sufficiently verified knowledge. Each stored element retains provenance information, including its source URL, source type, and verification timestamp.

\begin{figure}[!t]
\centering
\includegraphics[width=0.88\columnwidth]{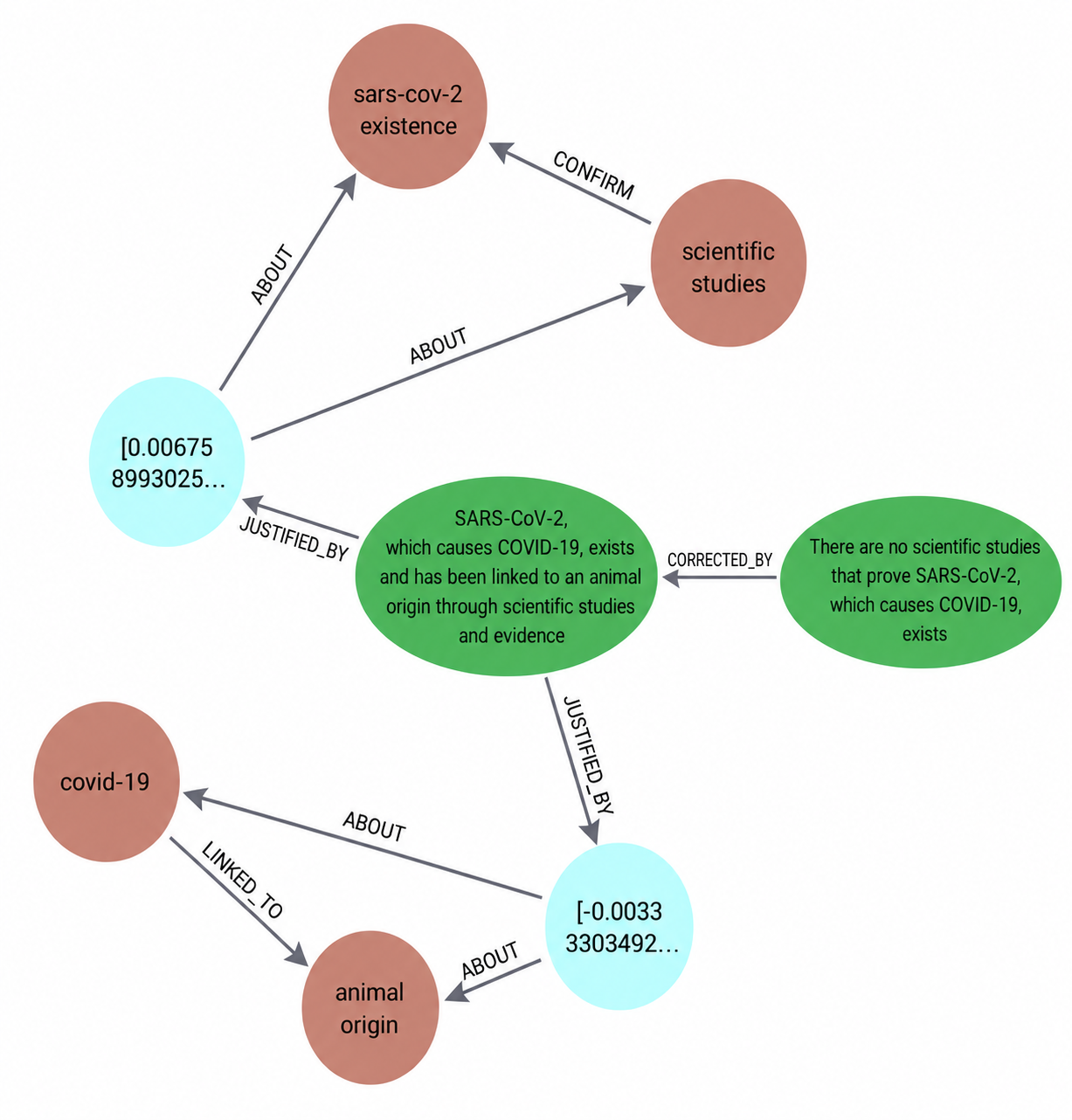}
\caption{KG persistence pattern for a false claim and its verified correction.}
\label{fig:kg-web-correction}
\end{figure}

The main graph relationships involved in this persistence process are illustrated in Fig.~\ref{fig:kg-web-correction}.

This closes the verification loop by converting externally verified
claims into reusable graph knowledge.

\section{Running Scenarios}

Table~\ref{tab:running_scenarios} summarizes four representative execution paths. In \textit{Scenario~1}, the claim \textit{``Asymptomatic COVID-19 patients can be infectious''} matches a stored claim with $S_{\text{SBERT}}=0.95$; NLI returns ENTAILMENT against a TRUE node, and the system resolves directly from the KG in ${\sim}$2s. In \textit{Scenario~2}, the claim \textit{``The level of symptomatic cases is greater than asymptomatic cases''} retrieves a topically similar but logically unrelated node ($S_{\text{SBERT}}=0.913$, NLI NEUTRAL with $p=0.958$); the high-neutral guard activates and routes to web investigation, preventing an unsupported KG verdict. In \textit{Scenario~3}, \textit{``Dexamethasone reduces mortality in hospitalized COVID-19 patients''} is resolved via the composite confidence score (Eq.~\ref{eq:confidence}), demonstrating graph-aware decision-making. In \textit{Scenario~4}, \textit{``Garlic cures COVID-19''} has no KG match; trusted web evidence contradicts the claim; the tribunal returns FALSE with corrected claim \textit{``Garlic does not cure COVID-19''}; triples are persisted for future reuse.
\begin{table}
\centering
\caption{Representative execution paths of the proposed system.}
\label{tab:running_scenarios}
\setlength{\tabcolsep}{3pt}
\begin{tabular}{|p{1.6cm}|p{4.2cm}|p{2.4cm}|}
\hline
\textbf{Scenario} & \textbf{Mechanism} & \textbf{Outcome} \\
\hline
Direct KG & MACRO retrieval + NLI entailment & TRUE from KG \\
\hline
High-Neutral Guard & High similarity, NLI neutral & Web fallback \\
\hline
Confidence scoring & SBERT + NLI + graph support & KG verdict accepted \\
\hline
Full web path & Tribunal + triple persistence & FALSE + KG update \\
\hline
\end{tabular}
\end{table}

\section{Evaluation}
\label{sec:evaluation}


Two COVID-19 claim collections were used to evaluate the proposed framework.

\textit{Dataset 1.}
The first evaluation set was derived from the COVID-19 fake-news
dataset introduced by Patwa \textit{et al.}~\cite{patwa2021}.
From the 6,420 labeled claims used in this study, 1,397 were reserved
for evaluation and the remaining 5,023 were used to construct the
initial Knowledge Graph. The evaluation claims were not inserted
into the graph, preventing direct overlap between KG memory and the
test set.

\textit{Dataset 2.}
The second evaluation set is a manually curated subset derived from the Check-COVID benchmark introduced by Wang \textit{et al.}~\cite{wang2023checkcovid}. From the publicly available repository, 333 self-contained claims were selected for evaluation. Claims requiring unavailable article context, examples without a definitive binary label, and duplicate or near-duplicate formulations were excluded. The original labels were mapped to the common \texttt{real}/\texttt{fake} label space used in this study. None of these claims were used during construction of the initial Knowledge Graph.


The standalone LLM baselines use
\texttt{meta-llama/Meta-Llama-3-8B-Instruct-Lite} and
\texttt{meta-llama/Llama-3.3-70B-Instruct-Turbo}, referred to as
\textit{Llama 3 8B Lite} and \textit{Llama 3.3 70B}, respectively.
Both baselines receive only the input claim and a fixed instruction
to return a structured factual verdict. They do not use KG retrieval,
web evidence, adversarial deliberation, or persistent memory.
Coverage denotes the proportion of claims assigned a definitive
\textsc{True} or \textsc{False} verdict; accuracy and macro-F1 are
computed over these resolved claims, while \textsc{Neutral} outputs
are treated as abstentions.

The resolution-path results are descriptive rather than controlled
ablations because the KG, tribunal, and recovery paths process
different claim subsets. Therefore, they characterize the behavior
of the complete framework but do not isolate the independent
contribution of each component.

\begin{table}
\centering
\caption{Performance on the curated COVID-19 test set (Dataset~2).}
\label{tab:main_results}
\setlength{\tabcolsep}{3pt}
\begin{tabular}{lrrr}
\toprule
\textbf{System} &
\textbf{Resolved} &
\textbf{Accuracy} &
\textbf{F1$_{\mathrm{macro}}$}
\tabularnewline
\midrule
Llama 3.2 8B &
333 (100\%) &
74.5\% &
74.4\%
\tabularnewline
Llama 3.3 70B &
324 (97.3\%) &
87.7\% &
86.3\%
\tabularnewline
\textbf{Proposed} &
\textbf{302 (90.7\%)} &
\textbf{97.4\%} &
\textbf{92.6\%}
\tabularnewline
\bottomrule
\end{tabular}
\end{table}

On Dataset~2 (Table~\ref{tab:main_results}), at 90.7\% coverage, the proposed system achieves 97.4\% accuracy on resolved claims, compared with 87.7\% accuracy at 97.3\% coverage for the Llama~3.3~70B baseline. The 31 unresolved (NEUTRAL) claims represent principled abstentions when neither KG nor web evidence meets confidence thresholds.

\begin{table}
\centering
\caption{Resolution path breakdown on Dataset~2.}
\label{tab:path_analysis}
\setlength{\tabcolsep}{4pt}
\begin{tabular}{lrrr}
\hline
\textbf{Path} & \textbf{Claims} & \textbf{\%} & \textbf{Acc.} \\
\hline
Knowledge Graph & 104 & 31.2\% & 93.3\% \\
Web -- Tribunal & 198 & 59.5\% & 99.5\% \\
NEUTRAL & 31 & 9.3\% & -- \\
\hline
\end{tabular}
\end{table}

Table~\ref{tab:path_analysis} shows the KG path resolves 31.2\% of claims at 93.3\% accuracy, while the tribunal resolves 59.5\% at 99.5\%. On Dataset~1 (Table~\ref{tab:noisy_results}), the system improves resolution rate from 59.6\% to 75.9\% and macro-F1 from 60.6\% to 69.1\% over the Llama~3.3~70B baseline, despite substantial dataset noise.
\begin{table}
\centering
\caption{Performance on the noisy Constraint English subset (Dataset~1).}
\label{tab:noisy_results}
\setlength{\tabcolsep}{4pt}
\begin{tabular}{lccc}
\hline
\textbf{System} & \textbf{Resolved} & \textbf{Acc.} & \textbf{F1$_{\text{macro}}$} \\
\hline
Llama 3.3 70B & 832 (59.6\%) & 80.2\% & 60.6\% \\
\textbf{Proposed} & \textbf{1061 (75.9\%)} & \textbf{80.6\%} & \textbf{69.1\%} \\
\hline
\end{tabular}
\end{table}

\textit{Error Analysis.} Among the 8 misclassified claims on Dataset~2, errors stem from: (a)~KG path matches with high lexical similarity but subtle qualifier differences (negation, scope); and (b)~web evidence reflecting outdated scientific guidance. During evaluation, 450 verified triples from 193 web-verified claims were persisted, demonstrating the continuous learning mechanism. KG-resolved claims complete in ${\sim}$2--3s; web-investigated claims require ${\sim}$50--60s, supporting the memory-first design rationale.

\section{Conclusion}

We presented a hybrid fact-checking system integrating Knowledge Graph semantic memory, NLI-augmented reasoning, trusted-source web retrieval, adversarial multi-agent deliberation, and persistent knowledge enrichment. The system achieves 97.4\% accuracy on resolved claims, outperforming standalone LLM baselines while providing evidence-grounded explanations and principled abstention on ambiguous claims. The continuous learning mechanism demonstrates that verified web evidence can be converted into reusable structured knowledge, progressively improving the system's internal memory. Future work will focus on temporal evidence filtering, negation-sensitive NLI models, ablation studies isolating individual component contributions, and cross-domain generalization beyond COVID-19 misinformation.


\vspace{0.3cm}
\textbf{Acknowledgment.}
This work was supported by a grant of the Ministry of Research, Innovation and Digitization, CCCDI-UEFISCDI, project number
PN-IV-P6-6.3-SOL-2024-2-0312 within PNCDI IV.

\bibliographystyle{IEEEtran}
\bibliography{bib}

\end{document}